\documentclass[journal]{IEEEtran}

\usepackage[utf8]{inputenc}
\usepackage[T1]{fontenc}
\usepackage{subcaption}
\usepackage{todonotes}
\usepackage{floatrow}
\usepackage{siunitx}
\usepackage{hyperref}
\usepackage{amsmath}
\usepackage{amsfonts}
\usepackage{booktabs}
\usepackage{pgf}
\usepackage{tikz}

\usepackage{soul} 

\begin{document}

\title{A Framework for Learning Predator-prey Agents from Simulation to Real World.}

\author{\IEEEauthorblockN{Jiunhan Chen, }
\IEEEauthorblockA{Department of Computer Science, Vrije Universiteit Amsterdam, the Netherlands\\}
\and
\IEEEauthorblockN{Zhenyu Gao, }
\IEEEauthorblockA{Department of Computer Science, Vrije Universiteit Amsterdam, the Netherlands}
}


\markboth{Journal of \LaTeX\ Class Files,~Vol.
}%
{Shell \MakeLowercase{\textit{et al.}}: Bare Demo of IEEEtran.cls for IEEE Journals}

\maketitle

\begin{abstract}
In this paper, we propose an evolutionary predator-prey robot system which can be generally implemented from simulation to the real world. 
We design the closed-loop robot system with camera and infrared sensors as inputs of controller. Both the predators and prey are co-evolved by NeuroEvolution of Augmenting Topologies (NEAT) to learn the expected behaviours. We design a framework that integrate Gym of OpenAI, Robot Operating System (ROS), Gazebo. In such a framework, users only need to focus on algorithms without being worried about the detail of manipulating robots in both simulation and the real world.  Combining simulations, real-world evolution, and robustness analysis, it can be applied to develop the solutions for the predator-prey tasks.
For the convenience of users, the source code and videos of the simulated and real world are published on Github \footnote{\url{https://github.com/chenjiunhan/Predators\_and\_Prey/}\label{git_video}}.
\end{abstract}


%
\IEEEpeerreviewmaketitle

\section{Introduction}
Predator-prey is a classical pursuit-evasion problem. A typical scenario is that there are a predator and a prey in a square arena, the predator must catch the prey within a certain time. The game ends when the predator catches the prey or time is beyond the limitation. The application of the solution of predator-prey problem might be used for searching, rescuing, exploration...etc. If we extend the predator-prey problem as multiple predators to catch a prey. The collaboration among multiple agents strengthen the reliability and scalability of completing a task.

The training for robots in the real world can be a time-consuming and money expensive task. An alternative way can be that we train the robots in the simulation world, and then we transfer the well-trained robots to the real world. Also, in the real world, we can even train the trained robots for a shorter time. The simulation environment can be created by Gazebo \footnote{http://gazebosim.org/}, which is a widely used 3D real-time simulation physics engine. 

There are different methods to solve the predator and prey problem. An early work from Bryson and Baron considered the pursuit-evasion problem as a differential game, it means the solution of the pursuit-evasion problem can be solved analytically when the map and the pose of the predator and the prey are known \cite{ho1965differential}. Another the work was from Raboin et al., they invented an algorithm to solve partially observable pursuit-evasion problem with optimizing the uncertainty of the prey \cite{raboin2012generating}. In the meantime, some researchers tried to solve this problem with Evolutionary Algorithms for various scenarios. We followed their steps and tried to explore interesting things from that point.

Our research topics includes that we implement different experimental settings to evolve the predators and show their advantages and disadvantages. Also, we aim to provide an simulated experimental environment for the predator-prey task.
From an experimental point of view, many aspects should be considered in this task. We can ask ourselves: "What's the number of predators? Can we change the number dynamically? Predators are controlled by a single controller or multiple controllers? Do we evolve the predators and the prey at the same time or we use a fixed strategy for one party? Is the environment fully observable or partially observable?" These questions help us to develop our research.


we expand our work to a partially observable environment. Instead of using an overhead camera, the sensors from the robots are used. For the more powerful sensors, in this paper we use the wheeled robots, Robobo \footnote{\url{https://theroboboproject.com/en/}} that combines a simple mobile base with your smartphone to create the next generation of educational robots. The robot has short-range (20cm) IR sensors and a camera by carrying a mobile phone on itself. In order to reduce the reality gap, we must simulate the camera and IR sensors in Gazebo. After finishing the evolution in simulation, we can transfer the evolved agents into the real world to further evolve or examine the performance. Both the predators and the prey co-evolved, even though the "arms race" might be not easy to be triggered, which is a key point for competitive coevolution. Fortunately, "Hall of Fame" is a technique that can stabilize the competitive coevolution process. We choose to co-evolve the sensor-based predators with an all-knowing prey in the simulation world, and the sensor-based predators can be transferred to the real world immediately. In the simulation world, we can access the data which is hard to get from the real world to help the evolution process. An all-knowing prey might encourage predators to develop better performance. Also, the fitness function for the predators is computed by the distance between agents, which is hard to get in the real world.



\section{Related Work}
\label{sec:related}

Most of the existing studies in evolutionary robotics address the co-evolution of pursuit and evasion strategies. That is, both the predator(s) and the prey(s) evolve their behaviour.
One of the early works is from Nolfi and Foreano \cite{nolfi1998coevolving} who perform co-evolution of one predator and one prey to investigate the ``arms races" phenomenon in a real world setting. The predator and prey are placed in a square world 47cm $\times$ 47cm. Several researchers simulated the predator-prey environment with the software. The article from Yong and Miikkulainen \cite{yong2009coevolution} shows that three predators corporately catch a prey, which is achieved in a 100 $\times$ 100 toroidal grid simulated world. Haynes and Sen \cite{haynes1997co} put the prey and the predators in 30 $\times$ 30 grid simulated world. Using such simple settings for simulation helps the research on different algorithms and strategies for predators. However, if the goal is to have robots applied in the real world then more realistic simulation environments are needed and even that may not be enough because of the infamous reality gap \cite{jakobi1995noise}.

Concerning the behaviour of predators, we can distinguish two cases: the predators can be homogeneous or heterogeneous, that is, the predators have the same controller or not. The work from Bryant and Miikkulainen \cite{bryant2003neuroevolution} notes that homogeneous agents can have some advantages compared to heterogeneous ones in some cases, for example, when the size of the team changes, although Haynes and Sen \cite{haynes1997co} reports that heterogeneous agents are better in the symmetrical setting like predator-prey. Most of them select neural networks as controllers. Yong and Miikkulainen \cite{yong2009coevolution} set inputs as the offset between "predator and prey" and "predator and predators" and choose heterogeneous agents, which makes the controllers able to coordinate different predators. We choose to use homogeneous agents to make our system easily scalable, i.e., our system works with the different number of predators. With features selection, we can reduce the dimensions of weights in the neural network, also make predators show obvious behaviour of the collision avoidance, which is similar to bird flock avoid the collision from each other in swarm intelligence \cite{beni2004swarm}.
\par Predators need a good opponent for evolving a complex strategy to catch a skillful prey. There are different strategies for prey in the literature. Jeong and Lee design a fuzzy logic controller using genetic algorithm \cite{jeong1999evolving}. Nolfi and Foreano \cite{nolfi1998coevolving}, \cite{nolfi2012co} evolved both the predator and the prey, which can suffer from "arms races" and the "red queen effect" \cite{paredis1997coevolving}. But the state-of-the-art shows that it can be improved by Cooperative Multi-objective Evolutionary Algorithms \cite{antonio2018coevolutionary}. In the work from Yong and Miikkulainen \cite{yong2009coevolution}, the evasion strategy of prey is directly away from the nearest predator. In this work, we design design a 'smart' strategy based on the gradient of the Gaussian function, which will be introduced in our previous study \cite{lan2019simulated}. It is easy to implement and applicable in both the simulated world and the real world.


\section{Methodology}
\label{sec:methodology}

\subsection{Robots}

\begin{figure}[!ht]
\centering
\begin{subfigure}{.48\textwidth}
  \centering
  \includegraphics[width=.9\textwidth]{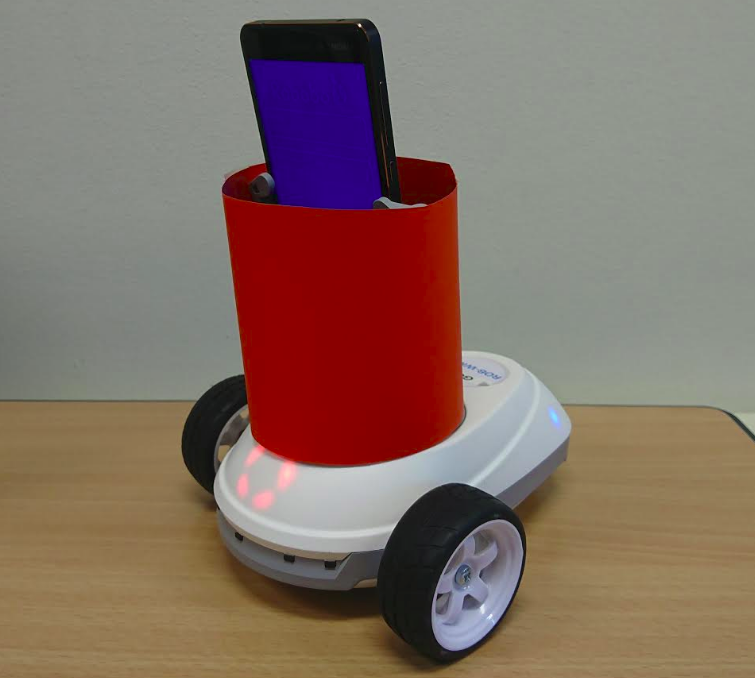}
  \caption{The real robot}
  \label{fig:robobo_real}
\end{subfigure}%
\begin{subfigure}{.49\textwidth}
  \centering
  \includegraphics[width=0.9\textwidth]{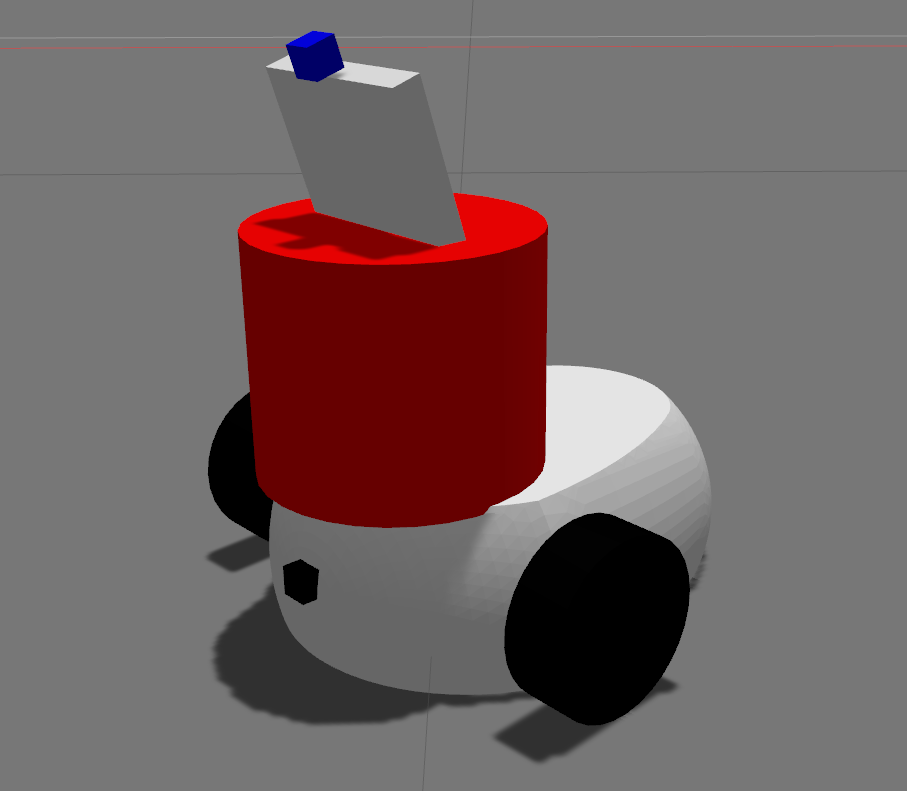}
  \caption{The simulated robot}
  \label{fig:robobo_simulation}
\end{subfigure}
\caption{Robobo in the real world and the simulation.}
\label{}
\end{figure}

In the paper, Robobo is our new robot to perform our experiments. Robobo is capable of carrying a mobile phone as a camera. We can use the camera to locate the relative position of the predators and the prey. We choose to use the front camera of the mobile phone to reduce the delay of data processing, because the resolution of the front camera is relatively low. Robobo has five front IR(infrared) sensors and three back IR sensors. The maximal range of IR sensors are only 20 cm. so we merely use one front IR sensor in the middle for collision detection. Robobo has five front IR(infrared) sensors and three back IR sensors, but we merely use one front IR sensor in the middle to avoid from collision. The blue cube in \autoref{fig:robobo_simulation} represents the camera of Robobo. The red cylinder is for the camera to detect the robot, and the prey can be distinguished by using green cylinder.

\subsection{The Simulation Environment}

\subsubsection{Integrating Gym of OpenAI, ROS, and Gazebo}

\begin{figure}[!ht]
\center
\includegraphics[width=1.0\textwidth]{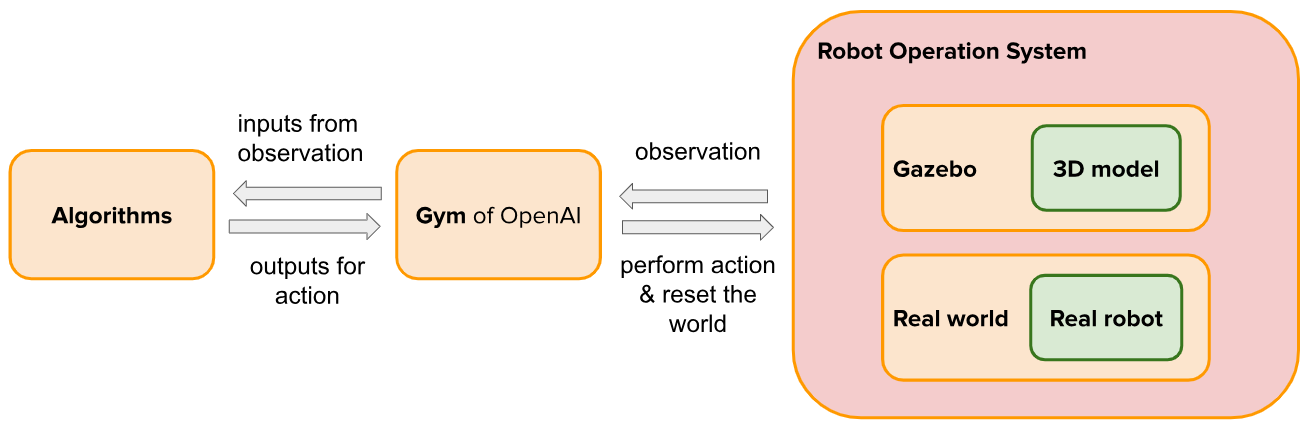}
\caption{Integrating Gym, ROS, and Gazebo. The users only need to focus on algorithms by receiving observed world data and generating output for action. The rest of things will be processed by Gym and ROS. RROS encapsulate Gazebo and the real world to provide the control interface for Gym.}
\label{fig:gym_ros_gazebo}
\end{figure}

\begin{figure}[!ht]
\center
\includegraphics[width=0.95\textwidth]{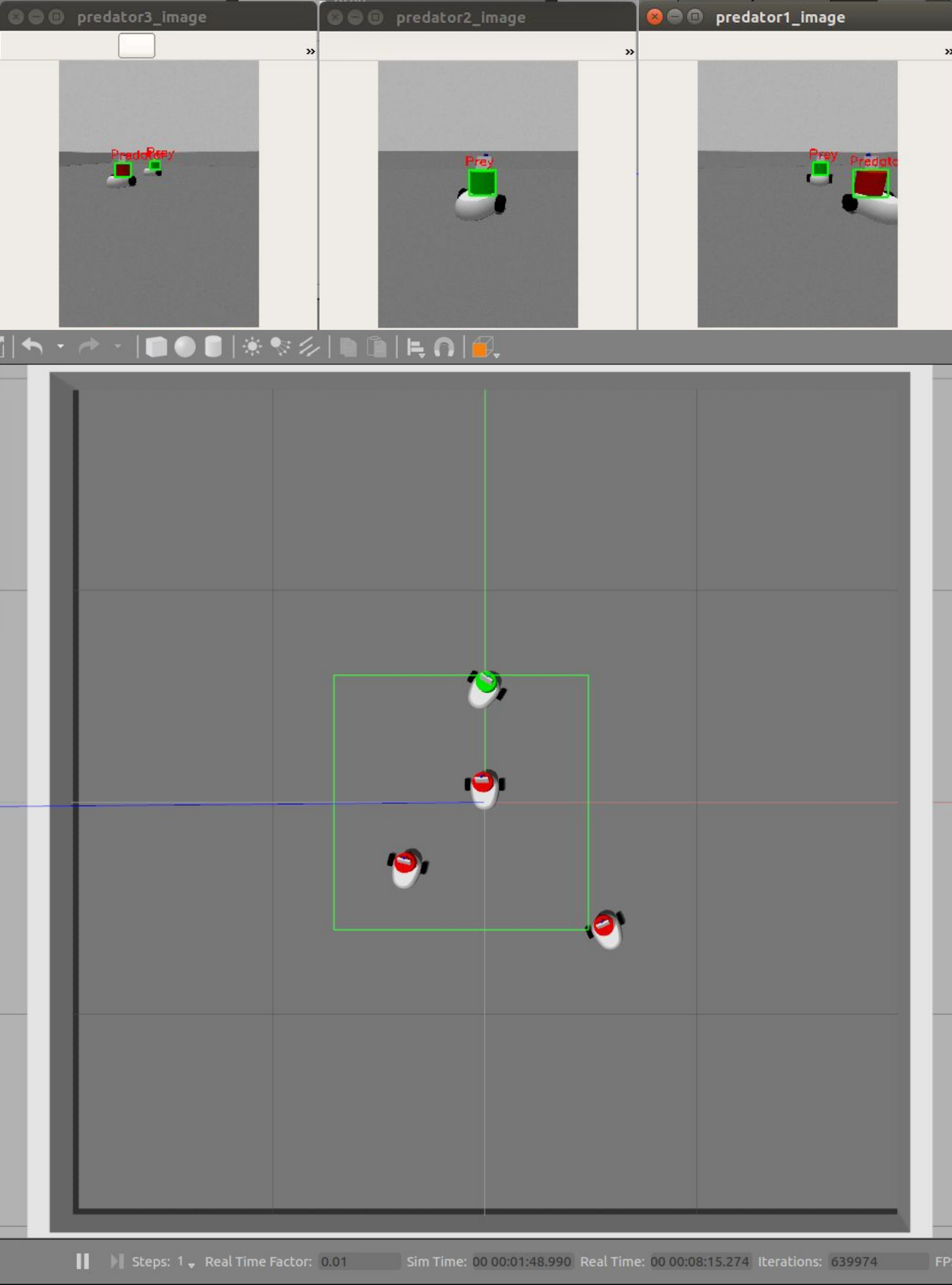}
\caption{The simulation environment. Three windows on the top are the views from different agents. The detected objects are selected by the bounding boxes.}
\label{fig:stage2_simulation}
\end{figure}

\begin{figure}[!ht]
\center
\includegraphics[width=0.95\textwidth]{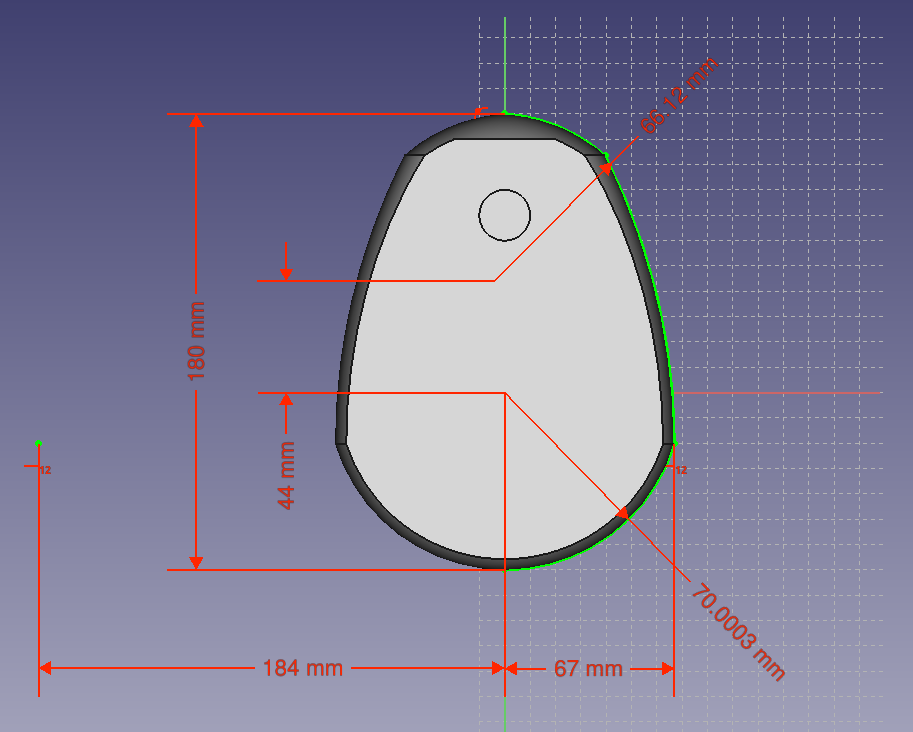}
\caption{Sketch of the robot in FreeCAD}
\label{fig:Sketch}
\end{figure}

Two of the most popular simulation environments are Gazebo and VRep. But it spends time to build an environment for research requirement. 
That's why there are platforms such as Gym\footnote{https://gym.openai.com/} from OpenAI to encapsulate the environment with an interface to perform action such that users can only focus on algorithms. So we integrate Gym, ROS, and Gazebo together to process all the details including controlling robots, resetting the experiment and getting world information. \autoref{fig:gym_ros_gazebo} shows the architecture of the integration.
\newline
\newline
In the beginning of one evaluation, the predators are placed parallel to one side of the wall, the prey is placed at the center.
The arena remains a square field, but the size is double as 4m$\times$4m to allow that the robots move more freely. \autoref{fig:stage2_simulation} shows the environment in the simulation world and the viewpoints from the different agents.
When the robot detects the other robots, a bounding box is created to select the region of detected agent by color detection based on hue difference among colors\cite{lan2018ICARCV}.
The gazebo allows user to simulate IR sensor and camera, so we try to make the difference between the simulation and the real world as small as possible.
The 3D model of the robots were created by FreeCAD, which is an open source technical drawing software to create 3D models.
\autoref{fig:Sketch} shows the sketch of the robot and its 3D model in FreeCAD.

\subsection{The Real World Environment}
\begin{figure}[!ht]
\center
\includegraphics[width=0.95\textwidth]{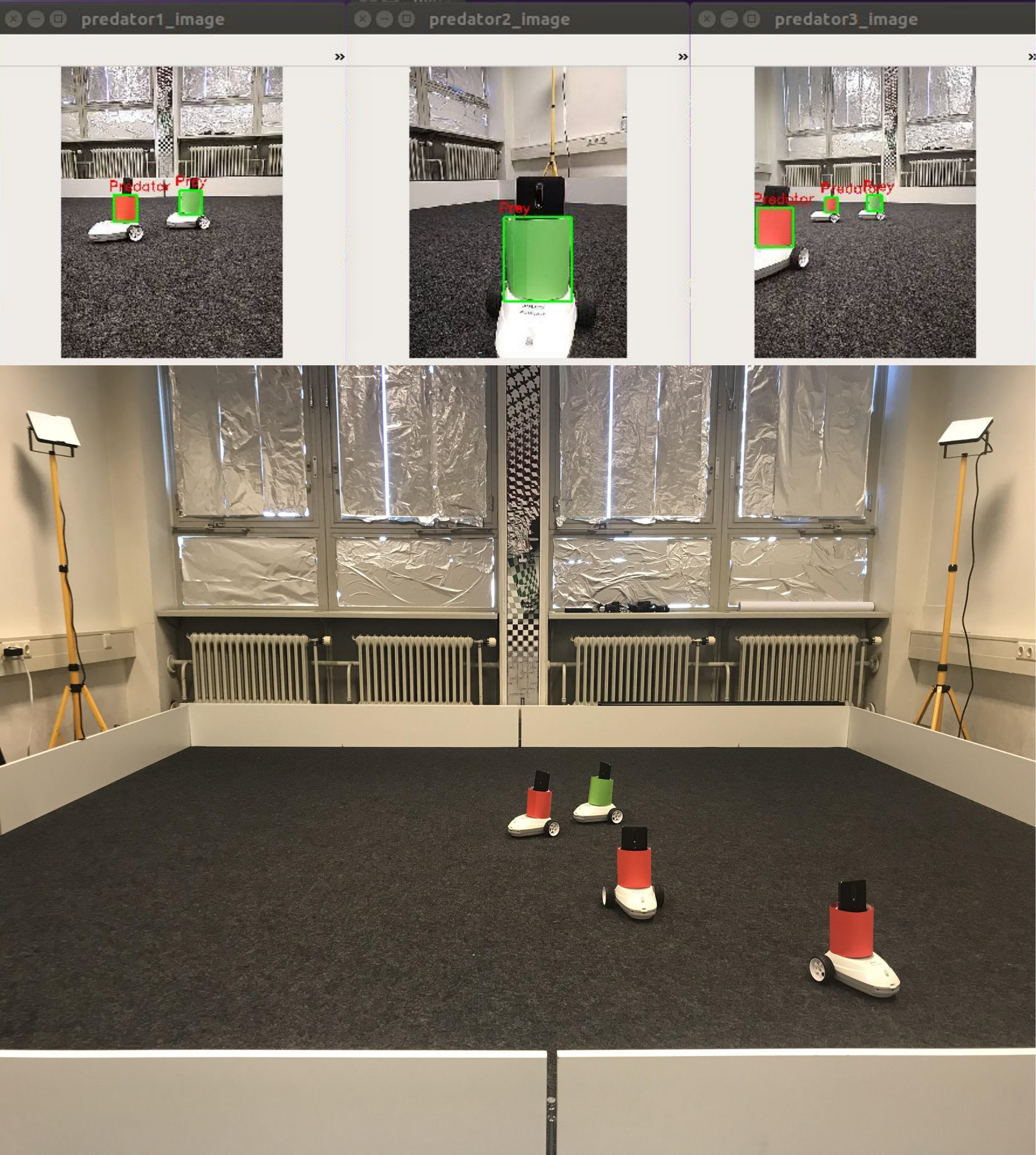}
\caption{The real world experiment. Three windows on the top are the views from different agents. The detected objects are selected by the bounding boxes.}
\label{fig:realall}
\end{figure}

The real world environment is a 4m$\times$4m arena. 
In this scenario, we use the embedded camera instead of overhead camera system.
The combination of camera and IR sensors are used to localize the coordinate of a robot itself and other robots. The real-time robot vision \cite{lan2018ICARCV,lan2019evolving} are partly applied to recognized other robots with colorful wrap. Three predators and one prey are connected with a computer via Wifi. A ROS master server is running on the computer. So robots pass their information from sensors to the computer via ROS, and then the controllers on the computer take data from sensors as inputs to generate outputs. In the final, the computer send control command back to the robots. \autoref{fig:realall} shows the viewpoints from the agents and the experimental setting in the real world.

\subsection{Partially Observable}
Unlike other systems with an overhead camera, we merely rely on the camera and the IR sensors on the robots themselves. This choice can extend the application of the predator-prey evolution system. However, we must face another challenge when we merely rely on the sensors of robots, which is that the world becomes partially observable for robots. When a robot is placed at a new place, it can't be sure where it is in this world and how it should move until the robot has detected the environment nearby itself. Take self-driving vehicles as an example, a self-driving system contains multiple deep neural networks to detect the environment, including the model to detect pedestrians, the model to detect traffic light...etc. Based on that environment information, the computer of the vehicle can make the decision of the next action. 

\subsection{Controller for prey}
The controller of the prey and the controllers for the predators are both evolved by NEAT. Our purpose is to evolve the predators with the sensors to catch the prey. One of the advantage of the simulation environment is that we can access specific data easily, compared to the real world. For example, we are able to directly get the pose of robots from Gazebo, however, in the real world, we must rely on some techniques like SLAM or GPS to locate the pose of robot. This fact allows us to create a all-knowing prey to help the evolution of predators. Here we list the inputs and the outputs of the prey:
\newline
\newline
\textbf{Input Layer:}
\begin{itemize}
\item{$\Delta \theta_1, \Delta \theta_2, \Delta \theta_3$: The angle difference between the orientation of the prey and the direction of the predators with relative to the prey.}
\item{$d_1, d_2, d_3$: The distance between prey and predators.}

\item{$x, y$: The coordinates of the prey.}

\end{itemize} 

\textbf{Output Layer:}
\begin{itemize}
\item{$\omega_L$: Angular velocity of the left wheel}
\item{$\omega_R$: Angular velocity of the right wheel}
\end{itemize}
Using hyperbolic tangent as the activation function.
\newline
\newline
To be aware of that the inputs are much more than the predators controller, which we can see in next subsection. This fact implies the mission is much harder for the prey, because the difference in the number of agents. Also, in most of situations, the predators only need to focus on moving toward the prey, however, the prey can't just simply run away from the predators, instead the prey has to turn 90 degrees to avoid from hitting the wall or being trapped from multiple predators. To evolve a better prey, we need a robot with wide-ranging sensors, otherwise, we have no choice but to use an all-knowing prey to make predators better.

\subsection{Heterogeneous controllers for predators}

Instead of the all-knowing prey, the predators can merely rely on their cameras and IR sensors. Using the heterogeneous controllers for each predator can increase the diversity of the predators and prevent from evolving into monotonous behaviour like only trailing the prey. We tried various inputs, but the simple inputs have a better effect. Here we list our inputs and outputs for the controllers of the predators:

\textbf{Input Layer:}
\begin{itemize}
\item{$x_{image}$: The horizontal position of prey shown in the image coordinate, the center is defined as zero.}

\item{$A$: The area of the prey in the image, and -1 if there is no prey in the image.}

\item{$c$: +1 if middle front IR sensor detects an object, otherwise -1.}    

\end{itemize} 

\textbf{Output Layer:}
\begin{itemize}
\item{$\omega_L$: Angular velocity of the left wheel}
\item{$\omega_R$: Angular velocity of the right wheel}
\end{itemize}
Using hyperbolic tangent as the activation function.

\subsection{Standard Coevolution Framework}\label{subsec:Standard Coevolution Framework}
\begin{figure*}[!ht]
\center
\includegraphics[width=0.99\textwidth]{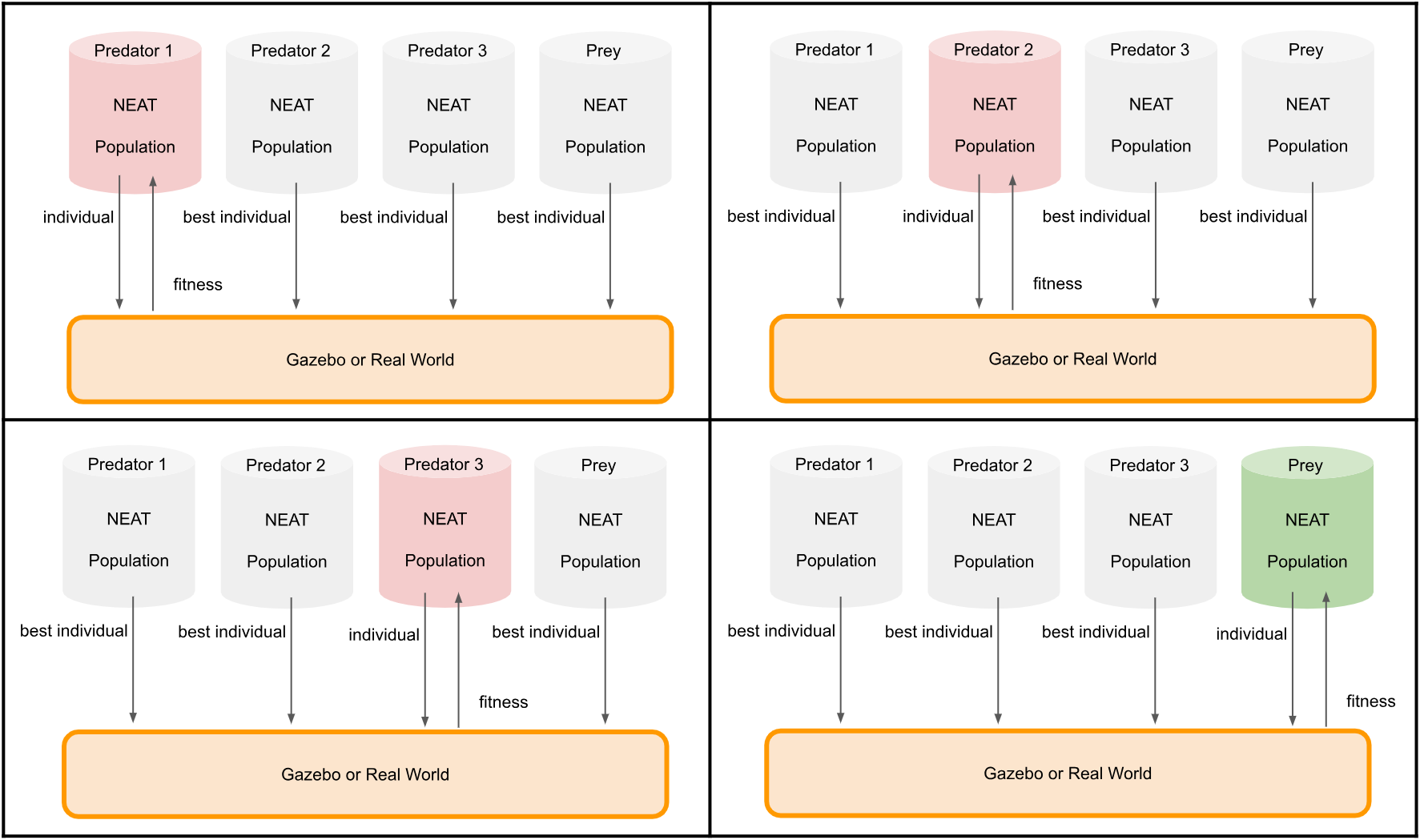}
\begin{subfigure}{.66\textwidth}
    \centering
    \includegraphics[width=0.99\textwidth]{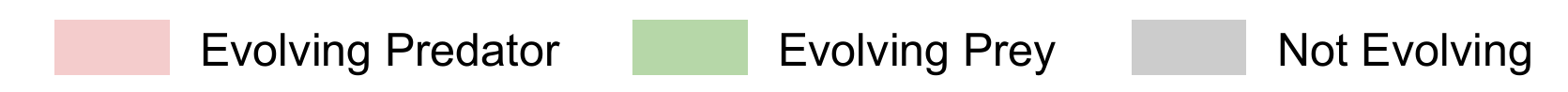}
\end{subfigure}
\caption{The standard co-evolution framework. Each agent evolves alternatively. The evolving agent(pink or green) tests its individual and get fitness as feedback. The other agents(grey) send the best controller from the previous generation to form the team or be the opponent(s).}
\label{fig:standard_coevolution_framework}
\end{figure*}

In the paper, we want to try heterogeneous controllers for the predators, therefore, there are four targets that need to be evolved, three predators and one prey. Assuming one of the predators is evolving, what are the teammates and the opponent that we should pick for an evolving target? For the teammates and the opponent we both pick the individuals with the best fitness from the previous generation, and randomly pick for initialization. It's more meaningful to cooperate with better teammates, because if we randomly pick two teammates for the evolved target, which leads to the performance of the team more depends on the teammates but not evolved target. The noise of fitness could be huge and more randomly. Also, it's more meaningful to compete with the best opponent, if the evolving predator can raise the performance even against the best opponent and the teammates are fixed comparing to the individuals in the same generation, which means the performance should be credited to the evolving predator. Otherwise, we don't know a good or bad performance should be credited to the evolving predator or a randomly chosen prey. Potter and De Jong talked about the architecture for the standard coevolution framework\cite{potter2000cooperative}. We apply the standard coevolution framework to our predators and prey task, and the architecture of coevolution can be drawn as \autoref{fig:standard_coevolution_framework}

we evolve all the agents alternatively. Except for the evolving target, the other agents are usually controlled by the best controllers from the previous generation. But we also implemented the technique "Hall of fame" for competitive coevolution, which means that the agent must play against the best opponents from each earlier generation. To reduce the evolution time, we make the evolving target to play against the best opponents from the randomly previous 10 generations. The parameters of NEAT is given in \autoref{table:Parameter of NEAT}

\begin{table}[!ht]
\center
    \begin{tabular}{ll}
    \toprule
    parameter    & value \\
    \midrule
    Number of generation   & 100  \\
    Population size & 20 \\
    Weight mutate rate  &  0.8 \\
    Bias mutate rate &  0.7 \\
    Probability of adding(deleting) connection &  0.1 \\
    Probability of adding(deleting) node &  0.1 \\
    Number of elites &  4 \\
    \bottomrule
    \end{tabular}
    \caption{The main parameters of NEAT}
    \label{table:Parameter of NEAT}
\end{table}

\subsection{Fitness Function}
We take the fitness function from \cite{rawal2010constructing} as a reference. For the prey, the fitness is computed by the survival time, the longer the prey survives, the higher the fitness can be. So the upper bound of fitness is when the prey never get caught during the evaluation time 30 seconds. For the predators, we choose to use selfish fitness function for each predator, in other words, only the final distance between evolved predator and the prey is considered. We want to emphasize that one evaluation is ended when one of the predators catches the prey. If there is an outstanding predator, it may lower the fitness of the other predators, which can motivate the other predators to become the first one who can catch the prey. But if all the predators are so selfish that they merely chase behind the prey, which may fail to catch the prey. We also tried to share the same fitness between all the predators, however, it seems that an outstanding predator usually makes the other predators lazy. The outstanding predator confuses others so that the other predators don't know if they are good. Here we display the mathematical representation of the fitness functions:
\newline
\newline
\textbf{Fitness function for prey:} \\
\begin{equation}
f_{prey}(\textbf{t}) = \frac{1}{K} \sum_{i=1}^K\frac{t_i}{T}
\end{equation}
T is a constant for evaluation time 30. $t_i$ is caught time at $i_{th}$ evaluation, $t_i = T$ if the prey is not caught. K is the number of evaluations. The opponents are selected by randomly previous 10 generations.
\newline
\newline
\textbf{Fitness function for a predator:} \\
\begin{equation}
f_{predator}(d) = \frac{1}{K} \sum_{j=1}^K\frac{1}{d_j}
\end{equation}
$d_j$ is the distance between the evolving predator and the prey at $j_{th}$ evaluation. K is the number of evaluations. The opponents are selected by randomly previous 10 generations. The reason that we use the inverse of distance instead of that a constant minus d is because there is a huge difference between "close" and "extremely close". Otherwise, the predators will be inclined to get easy points by staying somewhere close to the prey but never catch it.


\section{Results}
\label{sec:results}

The experiments include the evolution process and the evaluation in the simulation world, and evaluation in the real world. 

\subsection{Evolution and Evaluation in simulation}
As we introduced the standard coevolution framework in \autoref{subsec:Standard Coevolution Framework}, we evolved both the predators and the prey alternatively with 100 generations and relatively small population size 20. In our previous study \cite{lan2019evolutionary,lan2019simulated}, the small population size, even 13 is enough to evolve the agents. After the evolution process is finished, we tried to evaluate the controllers with Master Tournament, which means that we select the controllers with the highest fitness in every generation, and then we make them play against each other. So there will be 100 $\times$ 100 evaluation times for all the 100 generations of both the predators and the prey.

\begin{figure}[!ht]
\center
\includegraphics[width=0.95\textwidth]{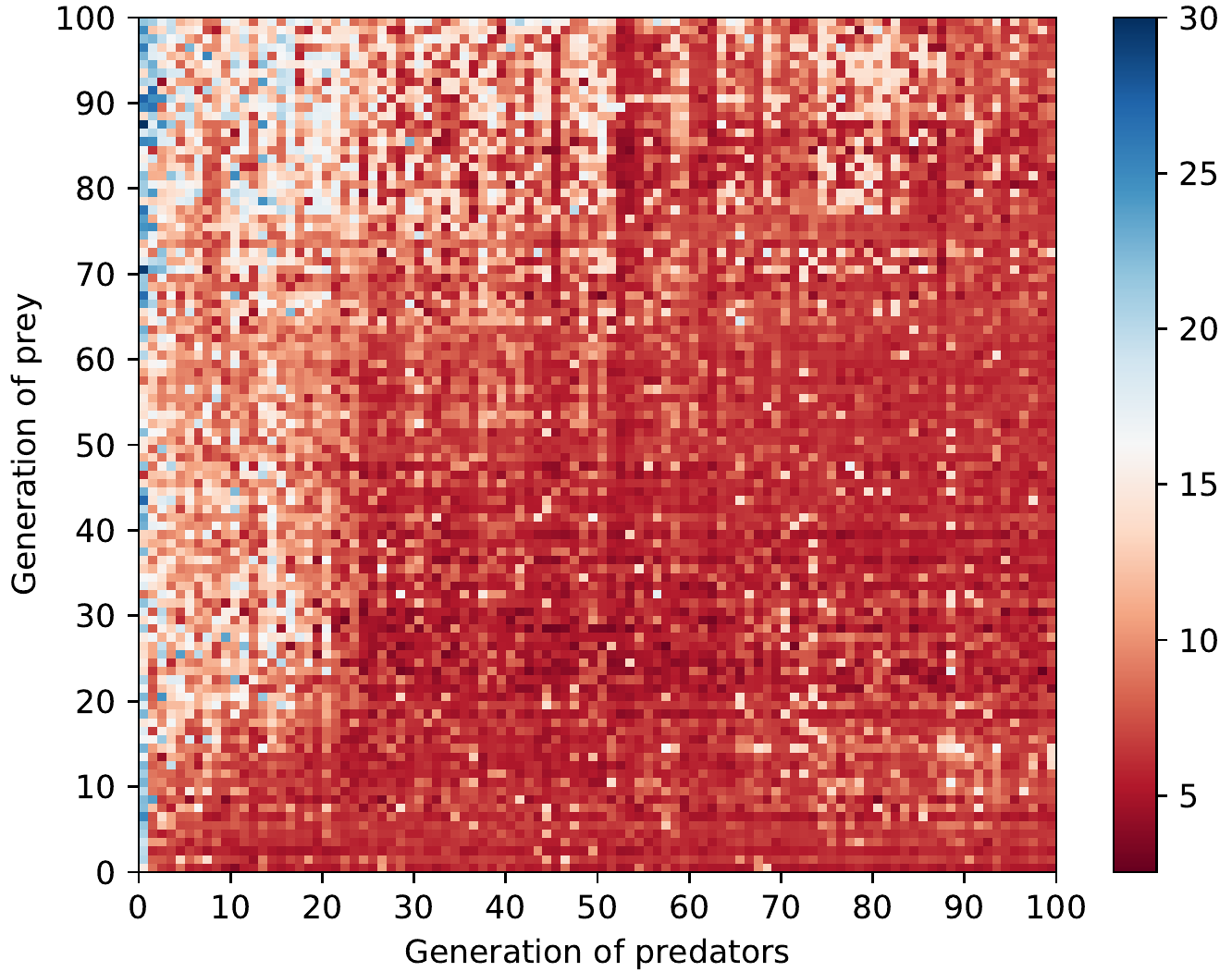}
\caption{The best prey from 100 generations versus the best predators from 100 generations. It shows a trend that the agent from the later generation performs better when it plays against the agent from the early generation.}
\label{fig:caught_time}
\end{figure}

\begin{figure}[!ht]
\center
\includegraphics[width=0.95\textwidth]{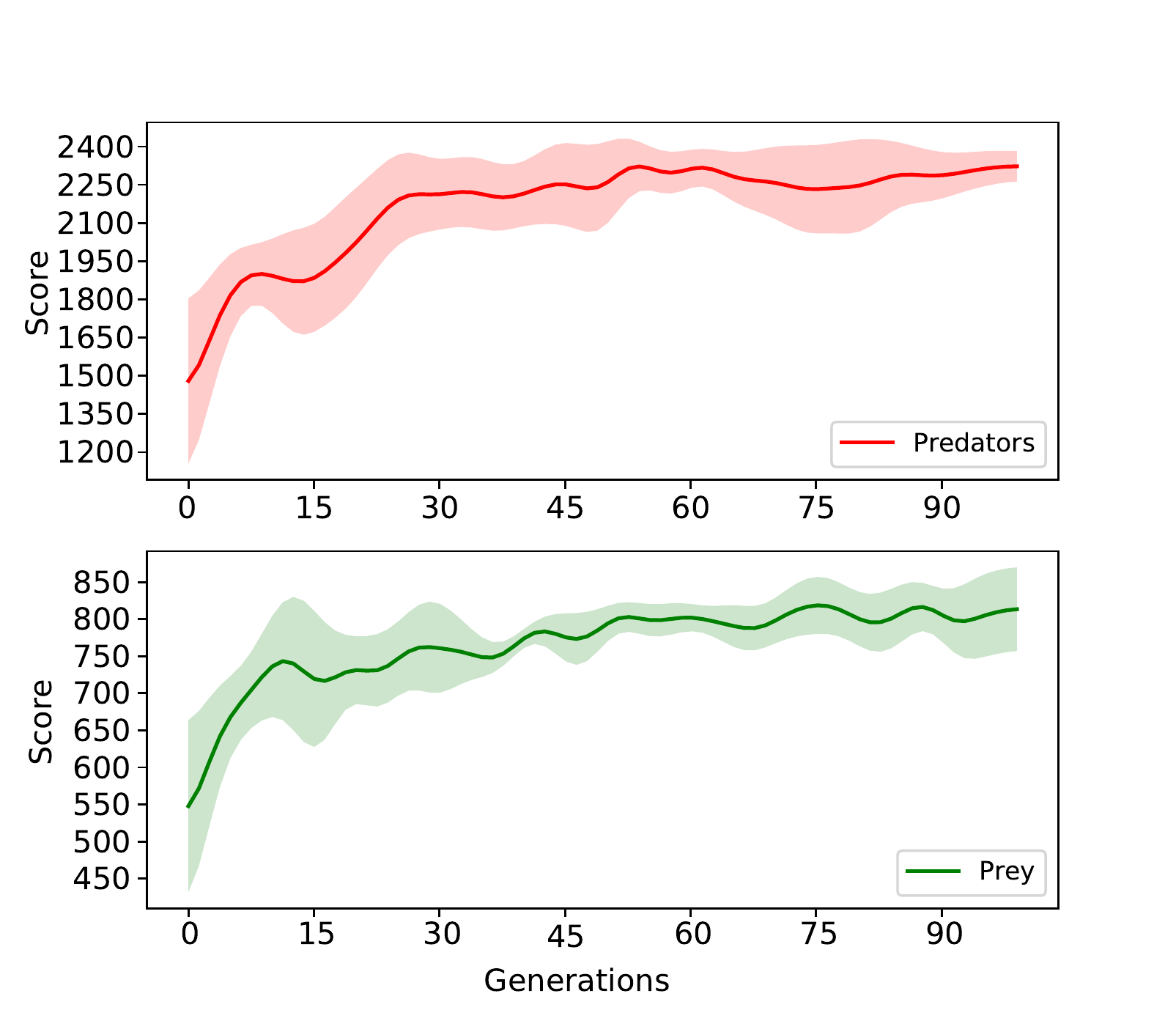}
\caption{The best prey from 100 generations versus the best predators from 100 generations. The graph shows the accumulated score for each generation and the final performance is stabilized. The shadow represents the range within one standard deviation. The line represents the average from three repeated experiment.}
\label{fig:accu}
\end{figure}

\autoref{fig:caught_time} shows the caught time of the master tournament. The later generation inclines to perform better when it competes against early generation. However, we can see that the task is more difficult to the prey from that most of values are below 10 seconds. Also the instability happened nearby 90th generation of predators and 13th generation of prey.

\autoref{fig:accu} is drawn by accumulated scores of agents. The score of prey is defined by caught time, and it accumulates a prey from a generation to play against all generations of predators. The score of predators is defined by that a constant minus the caught time, and the calculation of the accumulated score is similar. The graph shows that the coevolution can be stable in this evolution framework, although it seems that the upper limit of performance is hit around the 50th generation of predators.

\begin{figure}[!ht]
\center
\small
\begin{tabular}{c c c}
\hspace{-0.5cm} \includegraphics[width=0.34\textwidth]{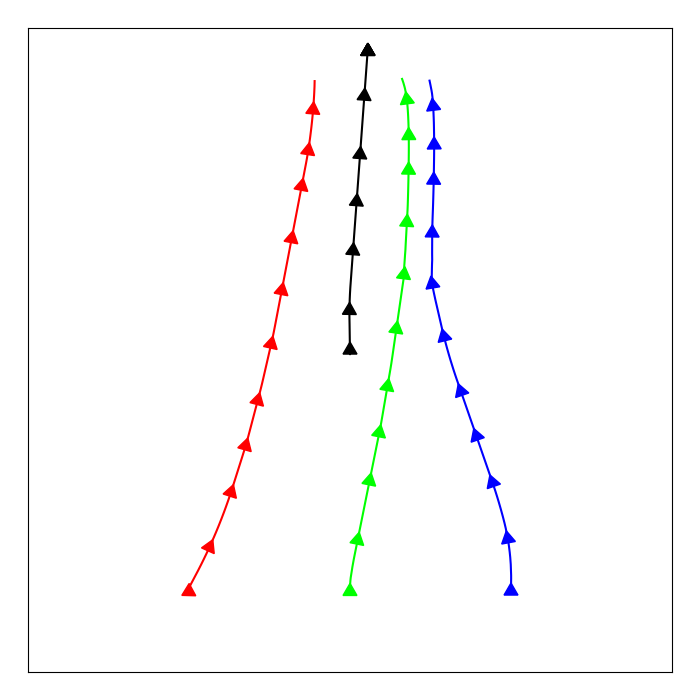} & \hspace{-0.6cm} \includegraphics[width=0.34\textwidth]{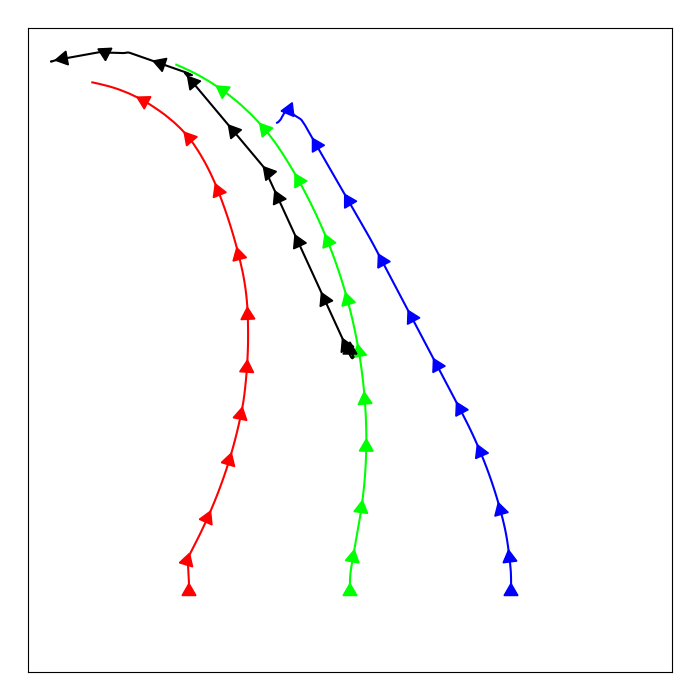} &
\hspace{-0.6cm} \includegraphics[width=0.34\textwidth]{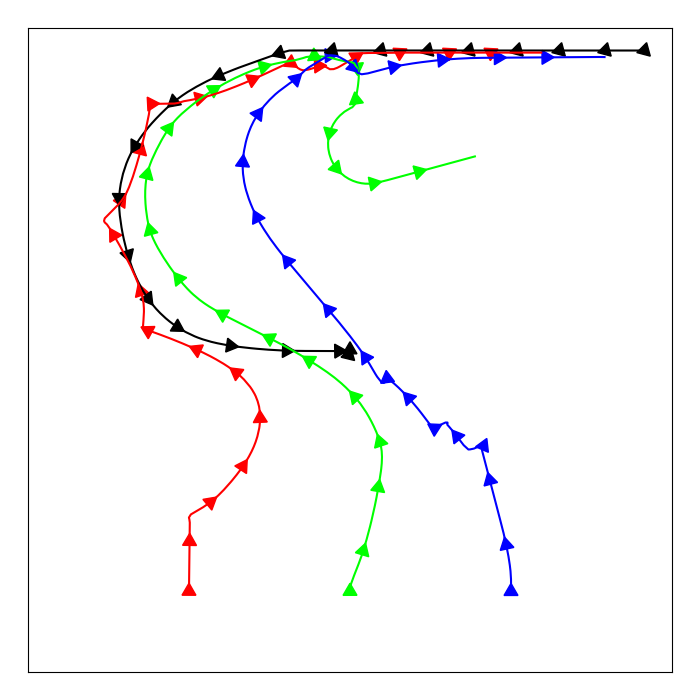} \\
\hspace{-0.4cm}(a) 0th generation & \hspace{-0.7cm} (b) 5th generation & \hspace{-0.7cm} (c) 8th generation \\
\hspace{-0.5cm} \includegraphics[width=0.34\textwidth]{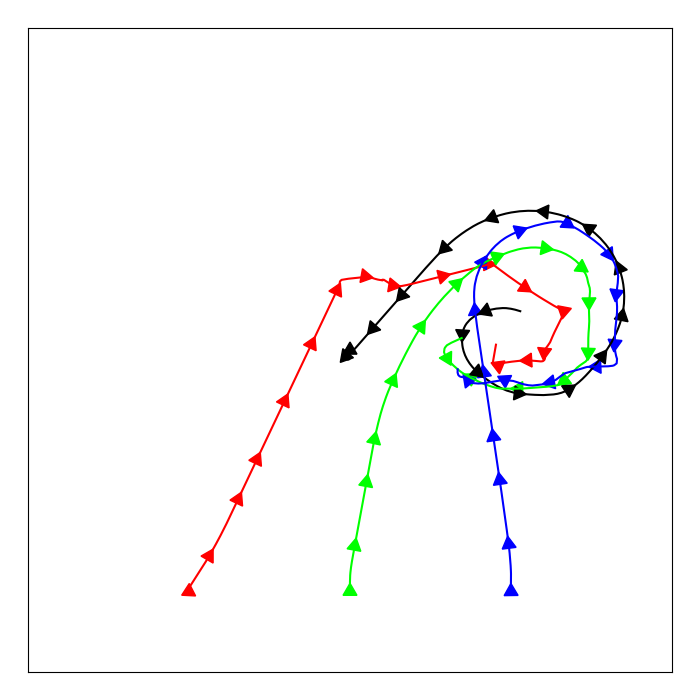} &
\hspace{-0.6cm} \includegraphics[width=0.34\textwidth]{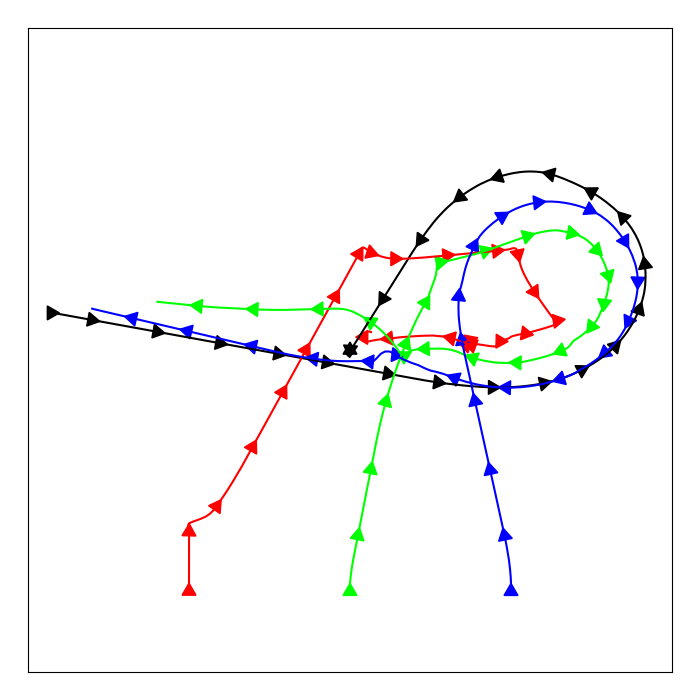} &  
\hspace{-0.6cm} \includegraphics[width=0.34\textwidth]{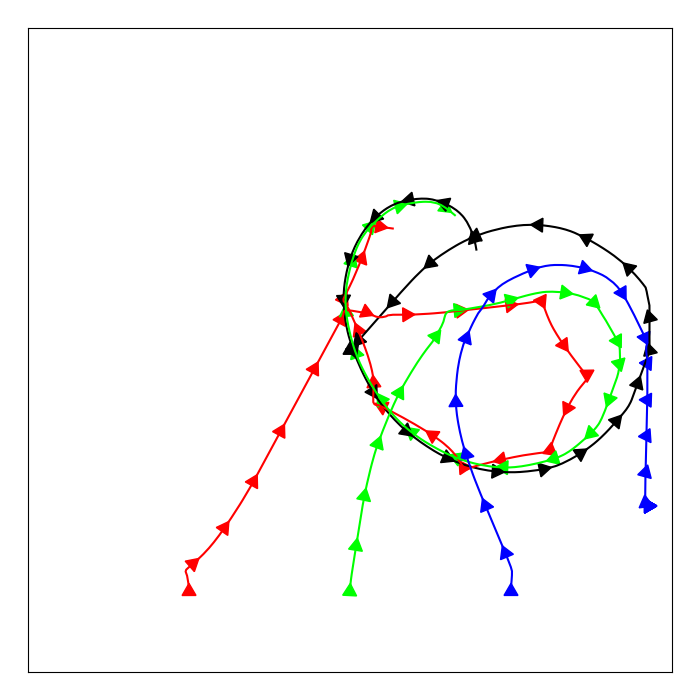} \\
\hspace{-0.5cm} (d) 20th generation & \hspace{-0.7cm} (d) 24th generation & \hspace{-0.6cm} (e) 41st generation\\
\hspace{-0.5cm} \includegraphics[width=0.34\textwidth]{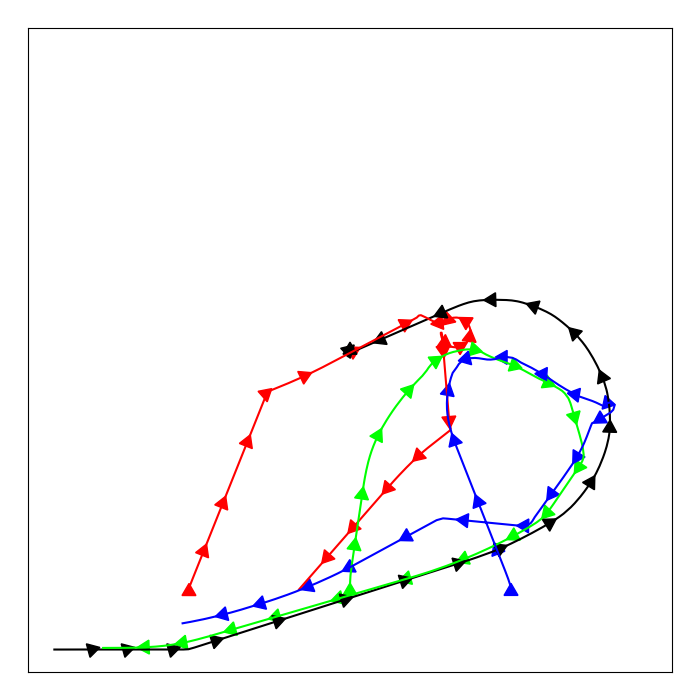} &  
\hspace{-0.6cm} \includegraphics[width=0.34\textwidth]{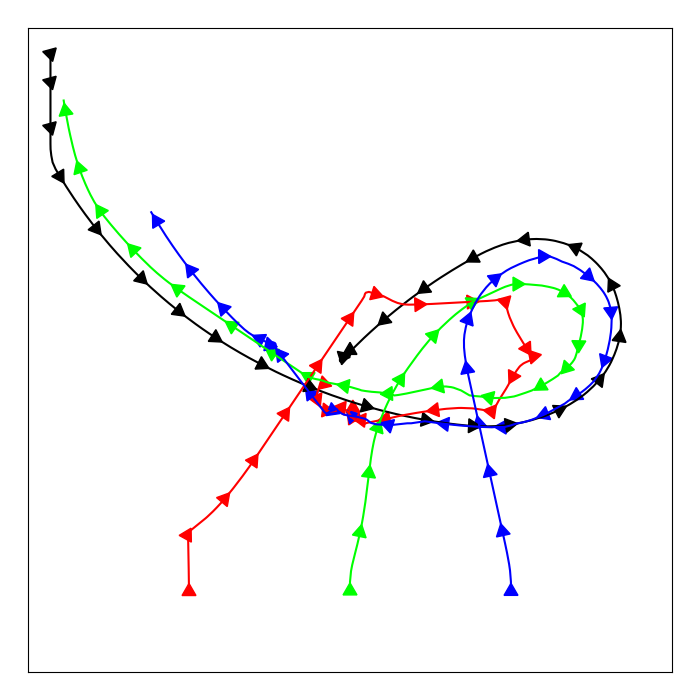} & \\
(f) 53rd generation & (g) 75th generation & \\
\end{tabular}
\caption{The trajectories of the predators from 41th generation versus the prey from the other generations}
\label{fig:generations_trajectories}
\end{figure}

To see that the behaviour was getting more complicated, we cherry-picked the predators from the 41st generation play against the preys from all the generations, and draw their trajectories as shown in \autoref{fig:generations_trajectories}. 
During the generations between 0 to 19, the best strategy for the prey is to move to the corner for prolonging the survival time. From the 20th to the 39th generation, the prey developed the strategy to slip away from the gap between the predator and the wall. The prey successfully went through the gap at the 24th generation. The generations from 40 to 100, Although, the performance for prey didn't change too much, the prey still tried to run away from the predators along different paths.

\subsection{Evaluation in the real world}

In the real world, the prey cannot be an all-knowing player, but we are still interested in evaluating the performance of the evolved predators when the controllers are transferred from the simulation world to the real world. So we use multiple human prey players to evaluate the performance of evolved predators in both simulation and the real world.

\begin{figure}[!ht]
\center
\includegraphics[width=0.8\textwidth]{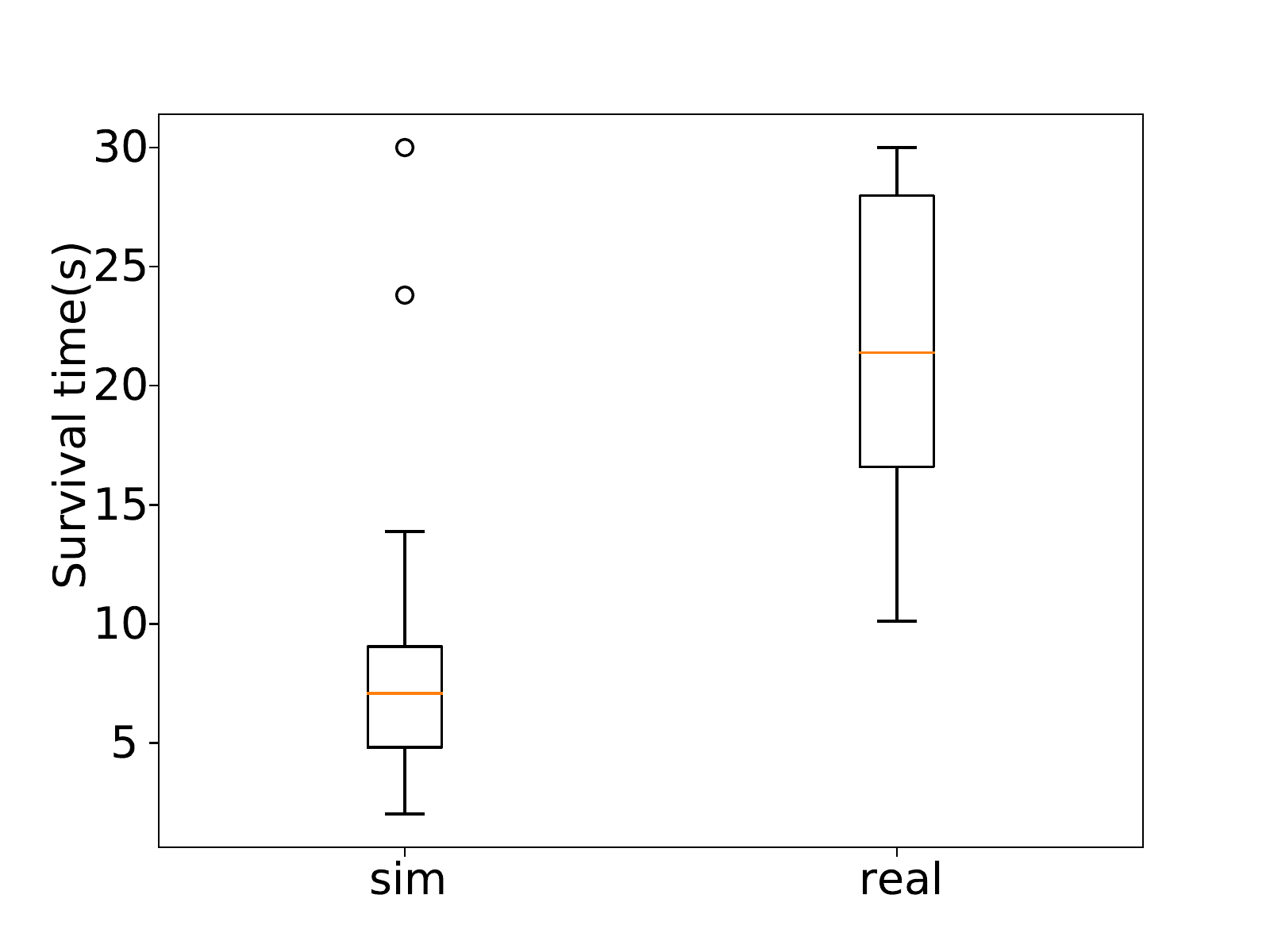}
\caption{The performance of the human controlled prey versus the evolved predators in both the simulation world and the real world.}
\label{fig:human_play}
\end{figure}

We created an interface for the human to control the predator with keyboard for both the simulation world and the real world. However the human players may introduce the bias, taking the average from multiple human players is necessary. The time spent to catch the human prey player is the criterion. The result can be shown in \autoref{fig:human_play}. The average survival time in the simulation world is 8.43 seconds, however in the real world the average survival time is 21.59 seconds. Even though the predators in the real world show the behaviour of pursuit, they still need further evolution to be applied in the real world. The reality gap will be further discussed in the \autoref{sec:discussion}.

\section{Discussion}
\label{sec:discussion}

\subsection{The reality gap}
\begin{figure}[!ht]
\center
\includegraphics[width=0.7\textwidth]{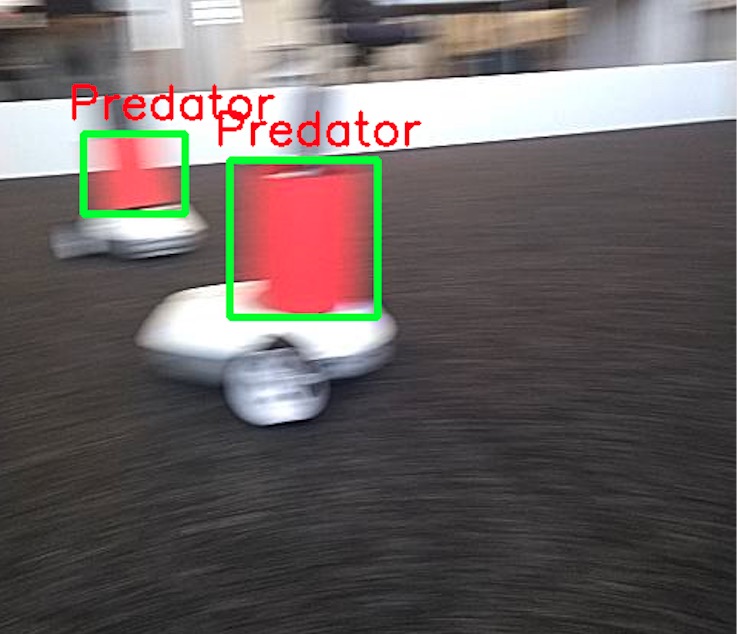}
\caption{The viewpoint from a rotating robot. Two predators are detected.}
\label{fig:rotation}
\end{figure}

The number of frames per second(FPS) could be a source which leads to increase the reality gap. FPS in the simulation world is around 10(because of the computation of simulation, image processing, and running algorithm), however due to the transmission of high quality of images via local WiFi, ROS and a router, the FPS in the real world can only be 5-7. The lower FPS has an impact on the object detection, some frames can be missed which leads to over rotate, even though during the rotation the object can be detected as shown in \autoref{fig:rotation}. 

\subsection{Triggering the arms race}
The arms race is not easy to be triggered. Even the observed behaviour becomes more complicate, it's not necessary that the performance is always better. However, the advantage of coevolution is to evolve various opponents without delicate human design. With the help of "Hall of Fame", the predators are able to catch various opponents and get higher average fitness. 
The variety of the evolution process makes the predators complete their evolution target in a certain degree. "Hall of Fame" can be implemented by choosing the opponent by random sampling. We also tried to select opponents from 10 previous generations, but it cannot stabilize the coevolution process. A sampling method for selecting opponents is necessary to stabilize coevolution. However, if we select opponent random sampling, it means that when the generation increases, it's possible that the sampling method, unfortunately, selects only bad performance controllers, which may lead to misestimating fitnesses, and it may further lead to hitting the upper limit of performance. In our experiment, the predators hit the upper limit of performance about 50th generation, so we may improve the performance with performance-based methods as mentioned in the work from Rosin and Belew \cite{rosin1997new}. 

\subsection{Fitness function}
In a few preliminary experiments, different fitness functions are tested. For the predator, fitness function $f(d) = constant - d$ and $f(d)=\frac{1}{d}$, $d$ is the distance between the prey and the evolving predator, have a huge difference, because it distinguishes the difference between "close" and "extremely close". The fitness function also can be defined by team performance, instead of only individual performance. In our other preliminary experiments, the fitness which defined by full team performance usually generates a lazy individual. It may be caused by that the difficulty of this task for multiple predators is relatively lower to the prey. The capitalism styled fitness function $f(d)=\frac{1}{d}$ works well at least in the simulation world, in other words, every predator works for itself to maximize the resource on its hands, but the whole team can be benefited from the result too. The predators are also not that selfish to just follow behind the prey, otherwise the severe collision should be observed, which means that they must collaborate in a certain degree.

\subsection{The Heterogeneous Controller}
The heterogeneous controller indeed shows the diversity compared to the homogeneous controller. According to our observation, some of the predators can rotate in a fixed direction(e.g. clockwise) to search the prey, and move toward prey directly when they find the prey, another observed controller can follow the prey smoothly and rotate in different directions which depends on where the prey disappeared in the views of predators. The various types of predators make they wouldn't just follow behind the prey as single homogeneous controller. 

\subsection{Evolution Time}
The processing of evolution takes a lot of computing time. There are two main reasons, the first one is because we need "Hall of Fame" to stabilize the process of coevolution, which makes our evolution n times longer than simple evolution, and n depends on how many previous generations to be competed with. Another reason is the time for simulating the camera and IR sensors to reduce the reality gap, it largely slows down the simulation speed of Gazebo. The whole evolution process needs 100 hours for 100 generations and population size as 20, which is around 300 times slower than using a homogeneous controller and a fully observable environment.

\section{Conclusion}

We presented a framework for evolutionary robotics in both simulation world and the real world by integrating Gym, Gazebo, and ROS. The interface provided by Gym allows users and researchers to focus on the learners, instead of wasting time on the setup with Gazebo in the simulation world or hardware in the real world. The framework can be used to investigate the tasks of cooperative coevolution and competitive evolution. Furthermore, reinforcement learning algorithms can be also applied as the leaner easily because our framework integrated the Gym package.

Our experiment shows that the standard coevolution framework can be also applied for sensor-based 3-versus-1 predator(s) and prey scenario. However, the transferability from the simulation to the real world can be limited by the hardware resource. The evolved robots still need to further evolve in the real world. The sampling for selecting opponents is the key step to stabilize the coevolution. At least the random sampling must be used. 

The processing of camera and IR sensors in simulation consumes large computational resource and therefore the evolution process takes a lot of computing time. 
We note that the requirement of computation resources enlarges the reality gap, because usage of computation resource and FPS cannot be consistent.

Although the framework of evolutionary robot works well from simulation to the real world, there are still many open issues to be further studied. In the future, we aim to apply a more effective learner for the evolution that can be done in a shorter computing time, for instance, the state-of-the-art data efficient learners, Bayesian optimization and Bayesian-Evolutionary Algorithm \cite{lan2020time}. This is always important issue for experiments in the real world. In addition, there are many other types of robots can be applied in this framework. We used the wheeled robots that is simple to be controlled but not so interesting. The tasks with other robots such as the modular robots with evolvable shapes \cite{lan2020learningA, lan2020learningB}, could be significant interesting. Last, the semantic new technology, knowledge graphs \cite{liu2020exploring,liu2020predicting}, can be applied for the human-computer interaction in future.

\ifCLASSOPTIONcaptionsoff
  \newpage
\fi

\bibliographystyle{IEEEtran}
\bibliography{bibliography}

\end{document}